\documentclass[10pt, a4paper]{article}

\usepackage[final]{lrec2026} % this is the new style
\usepackage{booktabs}
\usepackage{lipsum}
\usepackage{amsmath}

\usepackage{booktabs}
\usepackage{array}
\usepackage{tabularx}
\usepackage[table]{xcolor} % enables column shading
\usepackage{amsmath}       % for \boldsymbol
\newcolumntype{Y}{>{\centering\arraybackslash}X}
\newcolumntype{D}{>{\centering\arraybackslash\columncolor{black!6}}X} % shaded delta col

\usepackage{amssymb}   % for checkmark
\usepackage{pifont}    % for xmark
\usepackage{fontawesome5} % for warning icon (optional)

\newcommand{\cmark}{\checkmark}
\newcommand{\xmark}{\ding{55}} % consistent "X"
\newcommand{\warn}{\faExclamationTriangle}

\title{ConGA: Guidelines for Contextual Gender Annotation. A Framework for Annotating Gender in Machine Translation}

\name{Argentina Anna Rescigno\textsuperscript{1,2}, Eva Vanmassenhove\textsuperscript{3}, Johanna Monti\textsuperscript{2}} 

\address{\textsuperscript{1}University of Pisa, \textsuperscript{2}University of Naples ``L'Orientale'', \textsuperscript{3}Tilburg University \\
         argentina.rescigno@phd.unipi.it, e.o.j.vanmassenhove@tilburguniversity.edu, jmonti@unior.it\\
         }

\abstract{
Handling gender across languages remains a persistent challenge for Machine Translation (MT) and Large Language Models (LLMs), especially when translating from gender-neutral languages into morphologically gendered ones, such as English to Italian. English largely omits grammatical gender, while Italian requires explicit agreement across multiple grammatical categories. This asymmetry often leads MT systems to default to masculine forms, reinforcing bias and reducing translation accuracy. To address this issue, we present the \textit{Contextual Gender Annotation} (ConGA) framework, a linguistically grounded set of guidelines for word-level gender annotation. The scheme distinguishes between semantic gender in English through three tags, Masculine (M), Feminine (F), and Ambiguous (A), and grammatical gender realisation in Italian (Masculine (M), Feminine (F)), combined with entity-level identifiers for cross-sentence tracking. We apply ConGA to the gENder-IT dataset, creating a gold-standard resource for evaluating gender bias in translation. Our results reveal systematic masculine overuse and inconsistent feminine realisation, highlighting persistent limitations of current MT systems. By combining fine-grained linguistic annotation with quantitative evaluation, this work offers both a methodology and a benchmark for building more gender-aware and multilingual NLP systems.
 \\ \newline  }

\begin{document}

\maketitleabstract

\section{Introduction}
Detecting and mitigating gender bias in Machine Translation (MT) systems, and particularly in Large Language Models (LLMs), remains a central challenge in computational linguistics and research. Bias often arises from the structural tension between gender-neutral and gender-marked languages. 
While English can leave the gender unspecified, Italian requires explicit gender agreement across grammatical elements, compelling the system to select between masculine or feminine forms, even when the source text is neutral. For instance, the English sentence "\textit{I am a nurse}" may be translated into Italian as either "\textit{Io sono un'infermiera \texttt{<F>}}" or "\textit{Io sono un infermiere \texttt{<M>}}". Similarly, "\textit{I am happy}" becomes either "\textit{Io sono contenta \texttt{<F>}}" or "\textit{Io sono contento \texttt{<M>}}".

Previous research has shown that MT systems tend to overuse masculine forms due to unbalanced data, where male references are disproportionately represented \cite{cho_measuring_2019, prates_assessing_2020}. When the source language lacks gender cues, MT systems may resort to the most statistically likely or stereotypical option, often defaulting to masculine forms or reinforcing gendered associations (e.g., \textit{nurse} as female, \textit{engineer} as male) because of social biases \cite{monti_questioni_2017}. These biases are further exacerbated in cross-linguistic contexts, where the system must infer gender information absent from the source \cite{rescigno_case_2020, rescigno_gender_2023, vanmassenhove_gender_2024}.

Despite advances in neural architectures and context-aware translation, such biases remain unresolved \cite{savoldi_decade_2025}. Evaluating them also poses further significant challenges: quantitative methods rely on metrics such as precision, recall, or F1-score, while qualitative analysis explores context-rich examples, yet no unified framework exists to ensure comparability across languages or studies. Moreover, most existing research remains English-centric, neglecting low-resource and morphologically rich languages, as well as non-binary and gender-diverse identities.

The present study aims to address these gaps by introducing a linguistically informed framework for the annotation and evaluation of gender expression in translation. Specifically, it investigates how linguistic annotations can be leveraged to quantify bias in MT outputs.

By combining linguistic insight with computational evaluation, this approach aims to support fairer and more context-aware MT systems.

\section{Related Work}
Research on gender bias in MT highlights persistent asymmetries between gender-neutral and gender-marked languages. 
Early statistical MT systems were trained on large, often unbalanced, corpora, which have been shown to contribute to gender bias in their outputs \cite{stanovsky_evaluating_2019}. More recent neural MT systems have, in some cases, been shown to amplify gender-stereotypical translations by implicitly learning from such biased training data \cite{vanmassenhove_getting_2018, garg_word_2018, savoldi_gender_2021}.

Studies have also shown that when translating from English into morphologically richer languages, systems tend to over-generate masculine forms, even when the source is ambiguous \cite{vanmassenhove_lost_2019, rescigno_case_2020, rescigno_gender_2023}.

To mitigate these issues, several strategies have been explored. \citet{vanmassenhove_getting_2018} and \citet{elaraby_gender_2018} introduced gender tagging and speaker-aware MT to improve agreement, while \citet{escude_font_equalizing_2019} used debiased word embeddings. Furthermore, \citet{stanovsky_evaluating_2019} proposed the \textit{Target Gender Annotation}, marking tokens in the source with the grammatical gender of their aligned target translation to guide neural models. This method significantly improved accuracy on the WinoMT performance without lowering BLEU scores \cite{papineni-etal-2002-bleu}. Despite these advances, most strategies focus on the mitigation of bias at the model level, rather than the evaluation and annotation side of the problem.

\subsection{Evaluation and Annotation Frameworks}
Existing resources for the evaluation of gender bias in MT fall into two categories: benchmark datasets for evaluation and annotation schemes. 
Benchmarks, such as WinoGender \cite{rudinger_gender_2018} and WinoMT \cite{stanovsky_evaluating_2019} provide templated sentences to test coreference and expose occupational stereotypes, but they lack linguistic diversity. 
Datasets like MuST-SHE \cite{bentivogli_gender_2020} and gENder-IT \cite{vanmassenhove_gender-it_2021} extend the evaluation to speech and text translation, by providing contextual gender annotation. In MuST-SHE, gender information is defined at the segment level and derived either from the speaker's gender or from contextual lexical cues, allowing analysis of gender agreement and preservation in longer sentences. gENder-IT adapts this framework by offering explicit gender tags for systematic evaluation of gender realisation in MT.

More recent approaches emphasise contextual, linguistically grounded annotation. Instead of just counting the \textit{masculine vs. feminine} outputs, these approaches aim to describe how and where gender information is encoded in a text. \citet{hitti_proposed_2019} proposed a systematic taxonomy for gender bias, giving a definition of the phenomenon of \textit{gender generalisation}, among others. Building on this, \citet{havens_uncertainty_2022} expanded the classification to eleven categories, though mainly for English monolingual data.

These taxonomies emphasise inclusivity and context but are largely limited to monolingual English data and to detecting bias rather than measuring it quantitatively across languages.

While evaluation corpora for MT mostly focus on word- or sentence-level tagging, recent research dealing with large-scale models alignment introduces datasets for bias mitigation. GenderAlign \cite{zhang_genderalign_2024} presents a large dataset of dialogues annotated with unbiased or biased responses to fine-tune LLMs via Direct Preference Optimisation (DPO). It covers four different categories (such as stereotypes or discriminatory language) and shows that alignment on balanced distributions significantly reduces bias across the considered benchmarks. 

These alignment approaches differ from linguistic annotation schemes: rather than describing gender realisation in text, they train models to prefer non-biased outputs. Nevertheless, both fields converge on the need for transparent, well-annotated data and clear taxonomies to support reproducible bias evaluation.

\section{Theoretical Background}
Languages encode gender in different ways, using different linguistic elements. An \citet{EuropeanParliament2018GNL} report identifies three main types: 
\begin{enumerate}
    \item genderless languages (e.g., Finno-Ugric ones), which express the gender lexically (Finnish \textit{poika} "boy", \textit{tyttö} "girl") or by suffixes (in Hungarian, \textit{doctor} is the gender neutral form, and \textit{doktornő} is a "female doctor", with the \textit{-nő} suffix that means "woman"), and nouns and pronouns have no gender;
    \item natural gender languages (e.g., English, Swedish, Danish), which have mainly genderless nouns, but the pronouns are specific for each gender and reflect the referents’ biological sex;
    \item grammatical gender languages, which have gendered nouns/pronouns and an inflectional pattern for them (e.g., Slavic and Romance languages).
\end{enumerate}
The physical sex of animate entities, which is realised by the male/female polarity, varies across languages as it depends on the classification of reality, and is called \textit{\textbf{natural gender}}. However, some may argue that the concept of ``natural gender'' is a myth. Instead, \citet{corbett__2013} argues that English, and even many grammatical gender systems, reflect \textit{\textbf{notional gender}} \cite{nevalainen_its_1993}, culturally constructed ideas about sex and gender, rather than purely biological sex \cite{corbett__2013}.

\textit{\textbf{Grammatical gender}} is a morphological characteristic of languages, and can be masculine, feminine, or neuter. Usually, for inanimate entities, it is assigned by linguistic criteria; for animate entities, it will be coherent with the natural gender of the entity. Consequently, evaluating gender bias, which is inherently tied to the social and biological dimensions of human referents, is only meaningful for animate entities. Since inanimate concepts lack a natural gender, their grammatical gender assignment in the target language is a strictly morphological process. Therefore, it is impossible to establish a correspondence between grammatical and natural gender for inanimate objects, making them inapplicable for the evaluation of social or representational bias. \citet{corbett_gender_1991} also distinguishes the type of grammatical gender among languages, e.g., Italian (or German) has grammatical gender for nouns, pronouns, articles, and adjectives, while languages like English, which only have expressed gender in pronouns, are pronominal gender languages.

\subsection{Italian Grammatical Gender and Agreement}
Italian, like many Romance languages, uses grammatical gender (masculine or feminine), which is often assumed to reflect biological sex, with the consequent misconception that one can freely choose which grammatical gender to use when referring to animate entities. Grammatical gender has precise attribution and agreement rules, just like for singular and plural. 

Gender, but also number and person, is an inflectional category for Italian agreement. Inflection is a process of word formation in which words are modified to express different grammatical categories. 

Agreement occurs when a word changes form depending on the other words to which it relates. It is an instance of inflection, and usually involves making the value of some grammatical category (such as gender or person) "agree" between varied words or parts of the sentence. Agreement can be inherent, when the inherent traits of the word belong to the word \textit{per se}, such as gender for the names. Inherent inflection doesn’t require syntactic context \cite{booij_inherent_1996}, but the choice of the form is determined by what the speaker wants to express. For example, the noun \textit{attrice} (en: \textit{actress}) is inherently feminine as it describes an animate entity of female gender by definition.
On the other hand, contextual traits depend on the context in which the word is; therefore, the agreement is contextual and its inflection is dictated by syntax (e.g., agreement markers for adjectives); for example, an adjective can have a masculine or feminine form contextually to the noun it refers to, for instance, it: \textit{vecchia \texttt{<F>} insegnante} (en: \textit{old female teacher}) and \textit{vecchio \texttt{<M>} insegnante} (en: \textit{old male teacher}).
 
\section{Methodology}
For this experiment, we used the English sentences of the gENder-IT dataset \cite{vanmassenhove_gender-it_2021}, a bilingual English-Italian corpus derived from the MuST-SHE dataset (version 1.04) \cite{bentivogli_gender_2020}. 
The dataset originally provides word-level annotations for all nouns and pronouns referring to human referents for the English sentences. The set of tags created for annotating the gENder-IT dataset were \texttt{<F>} or \texttt{<M>} when it was clear from the sentence context that the referent was female/male, and a \texttt{<A>} tag when, within the given context, no assumption can or should be made with respect to the gender of the referent. We also performed a light cleaning of the dataset for the removal of the pre-existing tags. This was necessary to ensure that the tags were consistent with the proposed annotation guidelines.

For the Italian data, we used the gENder-IT dataset as an input to generate the translations. To do so, we used two large-scale models, mBART and TowerInstruct-7B model (\texttt{thinkverse/towerinstruct}, from now on referred to as TowerLLM), available via the Ollama platform\footnote{https://ollama.com/thinkverse/towerinstruct}. This model is part of the Tower family, which is designed for instruction-following tasks and optimised for high-quality text generation. The translations were produced using Ollama's local inference framework, allowing for on-device execution of LLMs. The TowerInstruct model was executed with a temperature setting of 0.2 to obtain more deterministic and consistent outputs, and also to ensure accuracy and semantic consistency in the translations, which are essential prerequisites for performing a reliable word-level alignment.
For this specific study, the primary goal is to build a reliable word-level gold standard for gender annotations. To achieve this, we had to isolate the baseline, therefore we maintained a temperature of 0.2. Higher temperatures, while introducing lexical variety, also increase the risk of hallucinations or semantic divergence from the source text, complicating the entity-level alignment required by the framework.
The annotation procedure was carried out using INCEpTION\footnote{https://inception-project.github.io} \cite{klie_inception_2018}, a tool that enables entity-level alignment between English and Italian sentences and is completely customisable to the experiment's needs.

The resulting parallel annotated corpus provides a fine-grained mapping between contextual gender cues in English and grammatical gender realisations in Italian, forming the gold-standard resource to be used for subsequent quantitative evaluation of bias through precision, recall, and F1 scores.

\section{Annotation Task}
While previous works have addressed gender bias in MT through evaluation of the outputs against reference translations or challenge sets \cite{prates_assessing_2020, stanovsky_evaluating_2019, vanmassenhove_getting_2018}, this approach targets an earlier phase, by defining exactly what types of gender-related phenomena contribute to evaluate reliably gender bias in both MT systems and LLMs.
The main aims of the task are:
\begin{itemize}
    \item to create an efficient framework for linguistic gender annotation to better detect bias in LLMs. 
    These annotations should clarify when and how gender is made explicit through agreement in the Italian language (e.g., adjectives, verbs, participles), and whether that realisation reflects a bias when derived from an otherwise ambiguous source;
    \item to annotate a dataset according to the framework described here, which will serve as a gold standard to assess biased orientation concerning the Italian language in an EN-IT machine translation task.
\end{itemize}
This task involves the annotation at the word level of the \iffalse semantic \fi grammatical and contextual gender of words in a bilingual setting (English-Italian).

\subsection{Task Explanation} 
Given the structural differences between English and Italian, the tagging approach was adapted to the specific role of each language in the translation process. In English (source), gender tagging is based on contextual meaning, as understanding gender cues in the source is essential for evaluating translation accuracy. In Italian (target), gender tagging relies on explicit grammatical realisations, such as agreement markers and gendered morphology, to identify how gender information is rendered when it is present or inferred. This distinction reflects a generalisable methodology for assessing gender in translation: capturing gender cues in the source to evaluate correctness, and analysing their realisations in the target to detect potential biases.
See Table \ref{tab:tag_example1} for an example.

\begin{table}[!ht]
\centering
\small
\renewcommand{\arraystretch}{1.2}
\begin{tabularx}{\columnwidth}{@{}lX@{}}
\toprule
\textbf{EN source} &
Women have been trained to think that \textbf{we} \texttt{<F1>} are overreacting or that \textbf{we} \texttt{<F1>} are being too sensitive or unreasonable. \\
\midrule
\textbf{IT target} &
Alle donne è stato insegnato a pensare che (noi) siamo troppo \textbf{reattive} \texttt{<F1>} o che (noi) siamo troppo sensibili o irragionevoli. \\
\bottomrule
\end{tabularx}
\caption{Example of tagging between English and Italian sentences. The tag \texttt{<F1>} highlights corresponding focus elements across languages.}
\label{tab:tag_example1}
\end{table}

In this example, the entity’s tag \texttt{<F>} for the English pronoun ``we'' results from the contextual agreement with the subject of the sentence, which is ``\textit{women}''. For the Italian translation, the tags would be applied to the adjectives, as elements bearing the gender information; in this specific case, then the \texttt{<F>} tag would go only to ``\textit{reattive}'', an explicit female form of the adjective, while ``\textit{sensibili}'' and ``\textit{irragionevoli}" are left untagged, as these adjectives have grammatically the same form for both genders, and considering the grammatical gender of these adjectives feminine would mean making an assumption, because as far as we know, this could be an analogous case to the example in Table \ref{tab:tag_example2}. Even though the subject is the same for both sentences (i.e., \textit{you} \texttt{<A1>}), the two adjectives referring to it have two different genders in the Italian translation (i.e., the adjectives \textit{asciutta} \texttt{<F>} / \textit{bravo} \texttt{<M>}).

\begin{table}[!ht]
\centering
\small
\renewcommand{\arraystretch}{1.2}
\begin{tabularx}{\columnwidth}{@{}lX@{}}
\toprule
\textbf{EN source} &
No, \textbf{you} \texttt{<A1>}’re still dry, \textbf{you} \texttt{<A1>}’re just being nice. \\
\midrule
\textbf{IT target} &
No, (tu) sei ancora \textbf{asciutta} \texttt{<F1>}, (tu) stai solo facendo il \textbf{bravo} \texttt{<M1>}. \\
\bottomrule
\end{tabularx}
\caption{Example of tagging between English and Italian sentences. Tags such as \texttt{<A1>}, \texttt{<F1>}, and \texttt{<M1>} indicate corresponding referents or gender features across languages.}
\label{tab:tag_example2}
\end{table}

\subsection{Tagging Strategy}
For the task, a minimal and functional set of \textbf{gender tags} has been defined. The tags reflect the semantic and grammatical realisation of gender in translation. For feminine and masculine animate entities, the tags used are \texttt{<F>} and \texttt{<M>}, respectively. For those cases where the gender cannot be said or inferred by any other element of the sentence, or can refer to a male or female entity, the tag \texttt{<A>} is used. However, the tags are applied differently across English and Italian due to structural differences between the languages.

In English, tags are assigned based on contextual cues. If the referent's gender is explicit or strongly implied (e.g., via context or coreference), it is tagged as \texttt{<F>} or \texttt{<M>}. Whenever the gender is not inferable (e.g., first-person pronouns like \textit{I}, \textit{you}, or gender-neutral roles like \textit{parent}), it is tagged as \texttt{<A>}.

In Italian, only \texttt{<F>} and \texttt{<M>} tags are used, and the annotation should reflect the grammatical gender realisations (e.g., adjective or participle agreement). We chose not to use any \texttt{<A>} tag in Italian, as the presence or absence of gender realisation is precisely what allows us to compute bias. Gender-neutral or ambiguous references that remain grammatically unresolved are left untagged.

This simplified tag set ensures that all gender realisations in Italian can be interpreted with respect to ambiguous or specified gender in English, enabling the construction of a bias score that reflects unjustified gender disambiguation in translation.

To support fine-grained analysis and the development of a bias evaluation metric, we enrich each tag with an \textbf{entity-level identifier} that distinguishes referents within the same sentence. As such, tags are enriched with an additional numerical tag that indicates which entities connect to each other and how many different distinct referents there are. Their annotation as \texttt{<A1>}, \texttt{<F1>}, \texttt{<M1>}, and so on, indicates distinct animate referents.

This indexing helps track how each ambiguous source entity is resolved in Italian. Moreover, this enables computation of how many ambiguous entities are present, and how many are realised with a gender in translation, providing a quantifiable basis for bias detection. For instance, if an English sentence includes two ambiguous entities, they would be annotated as \texttt{<A1>} and \texttt{<A2>}, respectively. The same criteria would be applied in the Italian sentence, with the tags corresponding to the gendered realisations of indexed English referents. If no gender is grammatically expressed, then no tag is applied. This neutrality is significant when assessing whether a gendered realisation was necessary.

\subsection{Specific Rules for Italian Annotation}
Italian annotations are restricted to cases where grammatical gender is explicitly realised. Only elements that morphologically encode gender are tagged with feminine or masculine labels, while ambiguous or unmarked forms remain untagged to highlight instances where gender is unnecessarily introduced in translation. Tags are applied to agreement targets such as adjectives, past participles with the verb \textit{essere} (to be), articles, possessives, and relative pronouns that display gender inflection.

Personal pronouns pose a challenge since Italian often omits the subject pronoun, as it is a pro-drop language; thus, tagging depends on agreement markers rather than the pronoun itself. Past participles agree in gender either with the subject (for intransitive verbs) or with the preceding object pronoun (for transitive verbs). 

Proper names are generally annotated as \texttt{<A>} in the English sentences, since names alone can not be considered reliable gender cues, given the cross-cultural naming variation and potential mismatch between a person's name and their pronoun preferences. An exception is made for cases where the full name of a well-known public figure is provided and their gender can be unambiguously verified through external sources (e.g., documented pronouns or biographical data). This approach follows the annotation criteria described in the gENder-IT corpus as well \cite{vanmassenhove_gender-it_2021}. However, in the Italian sentences, proper names are not directly tagged, but we tagged the gendered entity (article, adjective, past participle) that explicitly and directly refers to them. 

Finally, instances of overextended masculine forms are explicitly labelled with \texttt{<M>}, as masculine is not treated as a gender-neutral default.

\section{Annotation Procedure} 
The annotation was carried out manually on a sentence-by-sentence basis using the INCEpTION platform, which supports token-level tagging and entity alignment between the English and Italian texts. Each English sentence was first annotated to determine the contextual gender of all animate referents, following the tripartite tag set mentioned before. The corresponding Italian translations were then annotated by identifying the grammatical realisations of gender, such as agreement in adjectives, past participles, determiners, and pronouns. Ambiguous or unmarked forms were deliberately left untagged, as their neutrality provides essential evidence for gender omission or introduction in translation. 

The overview of the aggregated annotation statistics shows the number of \texttt{<M>}, \texttt{<F>}, and \texttt{<A>} tags obtained from the annotation. The total number of tags for the English sentences is 1559, significantly higher than the number of tags for the Italian sentences (see Table \ref{tab:tag_counts}). The main cause for that is that the majority of pronouns tagged in the source text are not present in the target sentences, as Italian is a pro-drop language; furthermore, English annotations also include the \texttt{<A>} tag, which was not considered for the Italian annotation task. 

To ensure consistency, the annotation followed detailed linguistic rules governing agreement, morphology, article-noun concordance, and overextended masculine forms. 
A trained linguist annotated the sentences for gender using a rule-based framework grounded in grammatical and contextual cues. The annotation followed explicit linguistic criteria, ensuring objective and consistent labelling across the dataset.

\begin{table}[!ht]
\centering
\small
\renewcommand{\arraystretch}{1.2}
\setlength{\tabcolsep}{5pt}
\begin{tabularx}{0.9\linewidth}{@{}lYYY@{}}
\toprule
\textbf{Gender Tags} 
& \textbf{EN} 
& \textbf{IT TowerLLM} 
& \textbf{IT mBART} \\
\midrule
\textbf{M}          & 356  & 544 & 570 \\
\textbf{F}          & 294  & 210 & 181 \\
\textbf{A}          & 909  & 0   & 0   \\
\textbf{Total Tags} & 1559 & 754 & 751 \\
\bottomrule
\end{tabularx}

\caption{Tag distribution by gender in English (EN) and Italian outputs from \textsc{TowerLLM} and \textsc{mBART}.}
\label{tab:tag_counts}
\end{table}

\section{Evaluation}
To quantitatively assess how faithfully gender information is preserved in translation, this study applies standard classification, such as Precision, Recall, and F1-score, to the annotated bilingual dataset. The comparison is carried out at the entity level, aligning tokens that share the same identifier across the two languages. The English annotations serve as the gold standard, while the Italian outputs produced by TowerLLM and mBART are treated as the models' predictions. 

For each gender, masculine or feminine, Precision is calculated as the ratio between the correctly generated gendered forms in the outputs and all gendered forms produced by the models.
\[
Precision = \frac{Correct\;Gendered\;Outputs}{Total\;Gender\;Tags\;(in\;Target)}
\]
Recall is the ratio between gendered entities in the outputs and all the gendered forms in the source text. 
\[
Recall = \frac{Correct\;Gendered\;Outputs}{Total\;Gender\;Tags\;(in\;Source)}
\]
F1-score, the harmonic mean of precision and recall, provides an overall index balancing accuracy and coverage. 
\[
F1 = 2 \times \frac{Precision \times Recall}{Precision + Recall}
\]

The evaluation metrics were computed automatically through a custom Python scrip designed to measure how faithfully each system preserved explicit gender tags and how ambiguities were resolved from source to target language. To achieve this, the script aligns the English source entities with the Italian target realisations based on their unique entity identifiers. 
The comparison is performed on a sentence-by-sentence basis across the datasets. For each aligned triplet (English source, TowerLLM, and mBART targets), the code categorises the translation into one of the three outcomes:
\begin{enumerate}
\item \textit{\textbf{matches}}, where the Italian translation reproduces the gender correctly as it is in the source;
\item \textit{\textbf{mismatches}}, whenever the gender differs from source to target texts; this was further distinguished between (a) \textit{error} and (b) \textit{bias}. We considered errors those cases where the gender tag in the target does not match with the gender tag in the source (i.e., M$\rightarrow$F or F$\rightarrow$M), while biases refer to the cases of mismatches when a gendered tag in the target sentence is used instead of an ambiguous entity in the source (i.e., A$\rightarrow$M or A$\rightarrow$F). In this latter case, translating an ambiguous source entity into a specific gender is not inherently an error and yields an acceptable sentence. However, it is exactly the arbitrary and asymmetric choice of one gender over the other that allows us to measure the model's bias;
\item \textit{\textbf{unmatched entities}}, regarding those entities that are present in the source but absent in the target or vice versa. 
%(Table \ref{tab:detailed-match-mismatch}).
\end{enumerate}

% ===================== TABLE: EXAMPLES BY CATEGORY =====================
\begin{table*}[!ht]
\centering
\small
\renewcommand{\arraystretch}{1.2}
\begin{tabularx}{\textwidth}{@{}lX@{}}
\toprule
\multicolumn{2}{@{}l@{}}{\textbf{\textsc{\cmark\ Match}} - Sentence 110 (F$\rightarrow$F)} \\
\midrule
\textbf{EN source} &
Women have been trained to think that \textbf{we} \texttt{<F1>} are overreacting or that \textbf{we’re} \texttt{<F1>} being too sensitive or unreasonable. \\
\textbf{IT TowerLLM} &
Alle donne è stato insegnato a pensare che \textbf{siamo} troppo \textbf{reattive} \texttt{<F1>} o che \textbf{siamo} troppo sensibili o irragionevoli. \\
\textbf{IT mBART} &
Le donne sono state addestrate a pensare che \textbf{siamo} troppo \textbf{reattive} \texttt{<F1>} o che \textbf{siamo} troppo sensibili o irragionevoli. \\
\addlinespace[0.6em]
\midrule

\multicolumn{2}{@{}l@{}}{\textbf{\textsc{\xmark\ Mismatch Error}} - Sentence 164 (F$\rightarrow$M)} \\
\midrule
\textbf{EN source} &
Many of the women working with me \texttt{<A1>} had to leave once \textbf{they} \texttt{<F2>} got married, because their husbands wouldn’t let \textbf{them} \texttt{<F2>} work. \\
\textbf{IT TowerLLM} &
Molte delle donne che lavoravano con me hanno dovuto lasciare il lavoro non appena si sono \textbf{sposate} \texttt{<F2>} perché i mariti non \textbf{li} \texttt{<M2>} volevano vedere lavorare. \\
\textbf{IT mBART} &
Molte delle donne che lavoravano con me dovevano andare via una volta \textbf{sposate} \texttt{<F2>} perché i loro mariti non \textbf{li} \texttt{<M2>} lasciavano lavorare. \\
\addlinespace[0.6em]
\midrule

\multicolumn{2}{@{}l@{}}{\textbf{\textsc{\xmark\ Mismatch Bias}} - Sentence 125 (A$\rightarrow$F/M)} \\
\midrule
\textbf{EN source} &
And he said, “No, \textbf{you} \texttt{<A1>}’re still dry, \textbf{you} \texttt{<A1>}’re just being nice.” \\
\textbf{IT TowerLLM} &
“No, sei ancora \textbf{asciutta} \texttt{<F1>}, stai solo facendo il \textbf{bravo} \texttt{<M1>}.” \\
\textbf{IT mBART} &
“No, sei ancora \textbf{secco} \texttt{<M1>}, stai solo facendo bene.” \\
\addlinespace[0.6em]
\midrule

\multicolumn{2}{@{}l@{}}{\textbf{\textsc{\warn\ Unmatch}} - Sentence 526 (A1$\rightarrow$unmatched ID)} \\
\midrule
\textbf{EN source} &
\textbf{Lindsay Malloy} \texttt{<A1>}: \textbf{They} \texttt{<A2>} told \textbf{Brendan} \texttt{<M3>} that honesty would “set \textbf{him} \texttt{<M3>} free,” but \textbf{they} \texttt{<A2>} were completely convinced of his guilt at that point. \\
\textbf{IT Tower} &
Lindsay Malloy: A Brendan dissero che l’onestà lo avrebbe “\textbf{liberato}” \texttt{<M3>}, ma erano completamente \textbf{convinti} \texttt{<M2>} della sua colpevolezza a quel punto. \\
\textbf{IT mBART} &
Lindsay Malloy: Hanno detto a Brendan che l’onestà lo avrebbe \textbf{liberato} \texttt{<M3>} ma erano completamente \textbf{convinti} \texttt{<M2>} della sua colpa a quel punto. \\
\bottomrule
\end{tabularx}
\caption{Examples of gender agreement across translation systems (\textsc{Tower}, \textsc{mBART}) for each evaluation category.}
\label{tab:examples_categories}
\end{table*}

Table \ref{tab:examples_categories} provides an example for each category and Table \ref{tab:detailed-match-mismatch} provides a detailed comparison of these three different categories identified after the tagging process.
On the basis of these comparisons, the script computes the metrics of Precision, Recall, and F1-score, separately for masculine and feminine realisations, treating the English annotations as the reference labels. 
These statistics are aggregated over all sentences to provide model-level measures for the accuracy of gender preservation (Table \ref{tab:detailed-match-mismatch}).

An additional component of the pipeline also addresses where the English source is ambiguous or unmarked (\texttt{<A>}). It analyses how each model resolves gender ambiguity by observing whether the Italian output adopts \textit{masculine} (A$\rightarrow$M) or \textit{feminine} (A$\rightarrow$F) forms. This step examines the models’ behaviour when gender marking is not explicitly present in the source, thereby quantifying potential gender bias in translation. Instances of A$\rightarrow$M or A$\rightarrow$F indicate that an ambiguous entity has been rendered with an explicit gender. While such realisations are not inherently erroneous, systematic asymmetries between them may signal the presence of gender bias.
The code additionally produces per-sentence logs and a .csv file reporting all matches, mismatches, and unresolved entities, facilitating a possible qualitative inspection. 

\begin{table}[!ht]
\centering
\small
\renewcommand{\arraystretch}{1.25}
\setlength{\tabcolsep}{4pt}
\rowcolors{2}{black!5}{white}

\begin{tabularx}{0.9\linewidth}{@{}lYY@{}}
\toprule
\ & \textbf{TowerLLM} & \textbf{mBART} \\
\midrule
\textbf{\cmark\ Match categories} & & \\
\hspace{1em}Match M & 173 & 165 \\
\hspace{1em}Match F & 104 & 88 \\
\textbf{Total Matches} & \textbf{277} & \textbf{253} \\
\midrule
\textbf{\xmark\ Mismatch categories} & & \\
\hspace{1em}Bias A$\rightarrow$M & 215 & 221 \\
\hspace{1em}Bias A$\rightarrow$F & 35 & 29 \\
\hspace{1em}Error M$\rightarrow$F & 0 & 2 \\
\hspace{1em}Error F$\rightarrow$M & 8 & 25 \\
\textbf{Total Mismatches} & \textbf{258} & \textbf{277} \\
\midrule
\textbf{\warn\ Unmatched EN IDs} & 488 & 493 \\
\bottomrule
\end{tabularx}

\caption{Detailed comparison of matching and mismatch categories for \textsc{TowerLLM} vs.\textsc{mBART}.}
\label{tab:detailed-match-mismatch}
\end{table}

\section{Results}
The metrics introduced in the previous section were applied to compare the performance of TowerLLM and mBART, using the manually annotated corpus as a reference standard. This section presents the results of our evaluation, focusing first on gender-wise accuracy and subsequently on the resolution of ambiguous cases.

\begin{table*}[ht]
\centering
\setlength{\tabcolsep}{6pt}
\small

\begin{tabular}{
    l
    >{\columncolor{black!10}}c
    c
    >{\columncolor{black!10}}c
    c
    >{\columncolor{black!10}}c
    c
}
\toprule
\textbf{Model} & \textbf{Match} & \textbf{Total Matches} & \textbf{Total Tags (EN)} & \textbf{Precision} & \textbf{Recall} & \textbf{F1-Score} \\
\midrule
\multicolumn{7}{c}{\textbf{Male}} \\
\midrule
\textbf{TowerLLM} & 173 & 544 & 570 & 31.8\% & 48.6\% & 38.3\% \\
\textbf{mBART}    & 165 & 570 & 356 & 28.9\% & 46.3\% & 35.5\% \\
\midrule
\multicolumn{7}{c}{\textbf{Female}} \\
\midrule
\textbf{TowerLLM} & 104 & 210 & 294 & 49.5\% & 35.4\% & 41.3\% \\
\textbf{mBART}    &  88 & 181 & 294 & 48.6\% & 29.9\% & 36.9\% \\
\bottomrule
\end{tabular}

\caption{Performance Metrics of Precision, Recall, and F1-score for \textsc{TowerLLM} and \textsc{mBART} by Gender. \textit{Alternating columns are shaded for readability}.}
\label{tab:metric_results}
\end{table*}

\subsection{Gender-wise Performance}
The gender-wise analysis evaluates how accurately each model reproduces explicit masculine or feminine references from the English source text in the Italian translation. Both models were assessed using the scores mentioned in the previous section. 

Table \ref{tab:metric_results} reports the Precision, Recall, and F1-scores of masculine and feminine outputs across the two systems. Overall, TowerLLM demonstrates slightly higher precision than mBART for both gender forms, indicating that its translations contain fewer erroneous outputs. However, mBART exhibits marginally lower recall, suggesting that it more frequently fails to reproduce gendered forms present in the source text. 

These results highlight a systemic masculine bias across both models: masculine gender forms are reproduced more frequently but with lower precision, whereas feminine gender forms are generated more selectively and often under-represented. This imbalance reflects a persistent male-default tendency in translation, where masculine forms serve as the default choice. Consequently, while TowerLLM offers modest improvements in accuracy, both systems exhibit asymmetric gender distribution in translation that reinforces existing gender biases in MT (Table \ref{tab:metric_results}). 

\subsection{Ambiguous Case Analysis}
Beyond explicit gender references, a crucial aspect of this evaluation concerns the translation of ambiguous entities: cases in which the source text provides no explicit lexical or contextual cues about the gender of an entity. Therefore, translating those entities into a morphologically richer target language compels the model to ``make a choice'', thereby exposing the gender bias.

The results in Table \ref{tab:detailed-match-mismatch} (see ``\textit{Mismatches}'') indicate that both models default to masculine realisations when ambiguity is present, showing high confidence in masculine predictions but very limited sensitivity to feminine alternatives. This male-defaulting tendency suggests that contextual cues in the English source are not sufficiently exploited to guide gender selection. The disproportionate number of A$\rightarrow$M (ambiguous-to-masculine) over A$\rightarrow$F (ambiguous-to-feminine) cases quantitatively confirms that gender introduction bias persists even in advanced LLM-based translation systems. This imbalance highlights a persistent structural asymmetry in how neutrality is resolved across languages, where masculine forms continue to function as the default. Moreover, the prevalence of A$\rightarrow$M translations suggests that models do not just lack information, but they actively apply a masculine-centric heuristic when in absence of source cues. 

\section{Discussion and Conclusions}
This study introduced the \textit{Contextual Gender Annotation} (ConGA) framework as a linguistically grounded methodology for detecting and evaluating gender bias in MT. By aligning semantic gender in English with grammatical gender realisation in Italian, our evaluation reveals a systematic pattern of asymmetry in how LLMs handle gender. Both TowerLLM and mBART consistently over-generate masculine forms and under-generate feminine ones, particularly when handling gender-neutral or ambiguous input. These findings confirm that gender bias persists even in advanced LLM-based translation systems, reflecting a systemic gender imbalance rather than random variation.

From a linguistic perspective, this pattern reinforces the notion of a ``\textit{male-default}'' translation bias observed in previous research \cite{cho_measuring_2019, vanmassenhove_gender_2024}. 
It also exemplifies what \citet{kahneman2011thinking} described as a \textit{systematic bias}, a consistent deviation that skews judgments in one direction. When such bias operates at scale, it poses fairness concerns, as repeated biased outputs may be reproduced or re-absorbed into future training cycles, amplifying gender imbalance over time.

Beyond the empirical findings, this work aims to address a major gap in the field, namely the lack of a standardised and scalable evaluation for gender bias in MT. While ConGA is primarily an evaluation framework, the annotated data it produces can directly support mitigation strategies, such as providing high-quality preference pairs for Direct Preference Optimisation (DPO) or fine-tuning models to improve gender-aware agreement.

By integrating linguistic human expertise with reproducible computational evaluation, ConGA offers a scalable approach for systematic bias assessment. The guidelines represent an important step in automatically annotated datasets. 
Finally, while we employed a low generation temperature to ensure high semantic consistency for our gold-standard annotation, we acknowledge that different sampling parameters might influence gender resolution. Exploring the impact of higher temperature settings on the stability of the male-default bias remains an important direction for future research.
Ultimately, linking linguistic insight and computational reproducibility makes the resulting dataset a reliable and reusable benchmark fro cross-system comparisons and broader cross-linguistic bias assessment. 

\section{Acknowledgements}
This work has been funded by the National PhD programme in Artificial Intelligence, partnered by the University of Pisa and the University of Naples “L’Orientale”, through the doctoral grant 39-411-24-DOT23A27WJ-7219 established by Ex DM 318, of type 4.1, co-financed by the National Recovery and Resilience Plan. Acknowledgements are also due to the support of The Eliza Centre for Humanities, Social Sciences, and Artificial Intelligence, whose interdisciplinary environment and resources have significantly contributed to the development of this research.

\section{Bibliographical References}\label{sec:reference}
\bibliographystyle{lrec2026-natbib}
\bibliography{lrec2026-example}

%\section{Language Resource References}
%\label{lr:ref}
%\bibliographystylelanguageresource{lrec2026-natbib}
%\bibliographylanguageresource{languageresource}

\end{document}